\documentclass[11pt,a4paper]{article}
\usepackage[hyperref]{acl2018}
\usepackage{amsmath}
\usepackage{amsfonts}
\usepackage{times}
\usepackage{color}
\usepackage{latexsym}
\usepackage{enumitem}
\usepackage{graphicx}
\usepackage{multirow}
\usepackage{booktabs}
\usepackage{bm}
\usepackage[linesnumbered,algoruled,boxed,lined]{algorithm2e}
\usepackage{url}

\aclfinalcopy 
\def\aclpaperid{181} 

\DeclareMathOperator*{\argmax}{arg\,max}

\definecolor{yamabuki}{RGB}{248, 181, 0}

\title{Event Extraction with Generative Adversarial Imitation Learning}
\author{Tongtao Zhang \and Heng Ji\\
Computer Science Department\\
Rensselaer Polytechnic Institute\\
\{zhangt13, jih\}@rpi.edu}

\date{}

\begin{document}
\maketitle
\begin{abstract}

We propose a new method for event extraction (EE) task based on an imitation learning framework, specifically, inverse reinforcement learning (IRL) via generative adversarial network (GAN). The GAN estimates proper rewards according to the difference between the actions committed by the expert (or ground truth) and the agent among complicated states in the environment. EE task benefits from these dynamic rewards because instances and labels yield to various extents of difficulty and the gains are expected to be diverse -- \textit{e.g.}, an ambiguous but correctly detected trigger or argument should receive high gains -- while the traditional RL models usually neglect such differences and pay equal attention on all instances. Moreover, our experiments also demonstrate that the proposed framework outperforms state-of-the-art methods, without explicit feature engineering.

\end{abstract}

\section{Introduction}
\label{sec:intro}


%
The event extraction (EE) task focuses on extracting structured event information (i.e., a structure of event trigger and arguments, ``\textit{what is happening}'', or ``\textit{who or what is involved}'') from unstructured texts.

In the past decade, many EE models and approaches have brought forth encouraging results by retrieving additional related text documents~\cite{ji2008refining,li2011exploiting,song2015light}, introducing rich features of multiple categories~\cite{hong2011using,li2013joint,zhang2017improving}, incorporating relevant information within context\cite{liao2010using,judea2016incremental} and adopting novel frameworks~\cite{chen2015event,feng2016language,nguyen2016joint,huang2016liberal}.

Most of supervised models seek the best mapping from features to labels based on the training documents.
However, there are still challenging cases: \textit{e.g.}, in the following sentences ``\textit{... the disobedience \textbf{campaigns} began last week.}'' and ``\textit{... Washington's anger with European resistance to the \textbf{campaign} was focused more on Paris}'' from two different documents on the similar topics on anti-war activities, the word \textit{\textbf{campaign}} can trigger either a \verb|Demonstrate| in the former sentence or an \verb|Attack| event in the latter.
%
With probabilistic approaches, the classifiers may prefer the category that appears more frequently in the training set.
Some methods may incorporate contextual information, \textit{e.g.}, the \verb|Arrest| events often co-occur with \verb|Demonstrate| events in the same document, but both documents mention \verb|Arrest|. 


%
Considering the process of how human annotators/readers understand the documents, a major difference between human methodology and these algorithms is that human do not always follow statistical results.
%
An incident with low frequency does not always imply that it is less important and human is able to reason on these corner cases if they benefit significantly more than other.
%
In order to simulate this process, a possible solution for those difficult cases is to assign more ``weights'', or \textbf{rewards} in terms of Reinforcement Learning as we utilize in our proposed framework.
 
\begin{table*}[htb]
\centering
\begin{tabular}{@{}lll@{}}
\toprule
  RL Terms  & Notations & EE Terms in supervised methods\\
  \midrule
  Agent & $A$ & Extractor\\
  Expert & $E$ & Ground truth from Human Annotators\\
  Policy & $\pi$ & Event model \\
Environment & $e$ & Data set, corpus, documents or shared features\\
State & $s$ & Features specified for a subtask\\
Action & $a$ & Labels: event types, argument roles\\
Reward & $r$ or $R$ & Loss functions, which stimulate the models to update\\
\bottomrule
\end{tabular}
\caption{A mapping table demonstrating the counterparts or equivalent concepts between reinforcement learning and EE approaches with supervised learning.}
\label{tab:term_mapping}
\end{table*}

Reinforcement Learning (RL) has been widely applied in Artificial Intelligence and robotics and recent applications of RL has defeated top human players in go~\cite{silver2016mastering}, shogi and chess\cite{silver2017mastering}.
Although these achievements are still far from ``Artificial General Intelligence'' or ``Strong AI'', it still reveals a promising direction of solving problems by emulating human.
Inspired from the previous successes, we model EE into a reinforcement learning problem.
The agent (or \textit{extractor} in this work) tags the entities and triggers, and detects the relation (argument roles) between the entities and triggers.
The extractor commits those actions with expectations of highest rewards according to its experience during training phase.

However, the original reinforcement learning methods are notoriously inefficient. The extractor requires huge amount of trials and errors especially when rewards are inappropriate, and reward shaping (enrich reward values with regard to states and actions) can be prohibitively expensive.
%
%
%
To tackle the problem, we adopt imitation learning -- specifically, inverse reinforcement learning -- to estimate the reward function, and apply the estimated reward function to the original reinforcement learning framework.
This framework manages to issue proper rewards to the extractor, especially for those challenging triggers and arguments.
The rewards are estimated by a Generative Adversarial Network (GAN) which takes the input of the expert (ground truth from annotators) and extractor as well as the states.
The GAN ensures the highest reward for the expert at a certain state and the extractor attempts to imitate the expert by pursuing the highest rewards.

The main contributions of this paper can be summarized as follows: 
\begin{enumerate}[noitemsep,nolistsep,leftmargin=*]
\item We apply reinforcement learning framework to event extraction and we demonstrate that a proper and dynamic reward function which is estimated from states and actions ensures more optimal performance in a complex RL task.
\item Without excessive feature engineering, our proposed framework outperform state-of-the-art approaches based on explicit feature extraction.
\item We also prove that shared parameters and embeddings across multiple EE subtasks in neural networks can improve the performance.
\end{enumerate}
\section{Task and Term Preliminaries}
\begin{figure*}[t]
\centering
\includegraphics[width=0.95\textwidth]{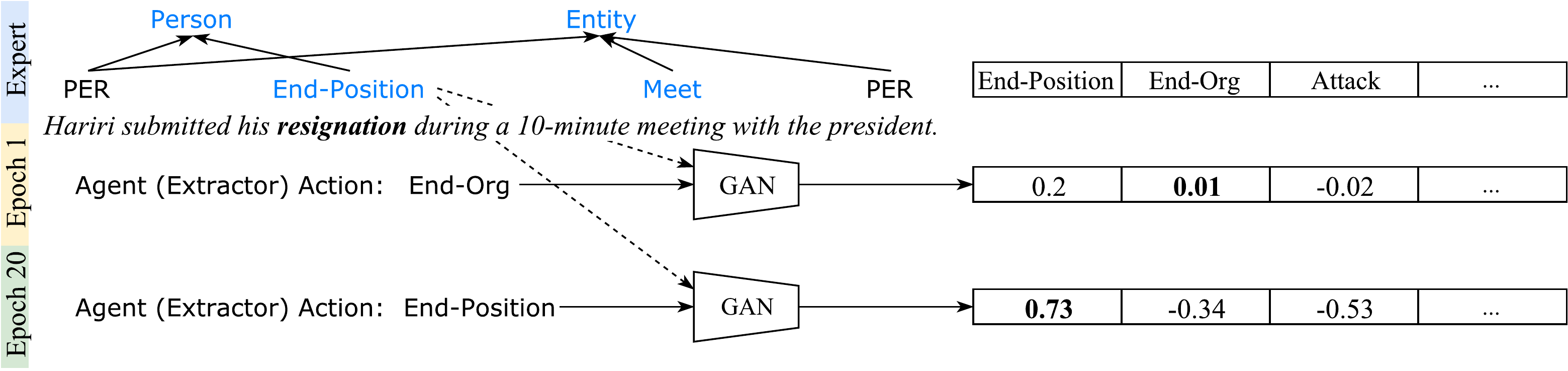}
\caption{A diagram for ACE event extraction task. The lower half denotes a process where the extractor selects the event type for trigger word ``\textit{resignation}'' at different epochs. Reward function values are estimated from generative adversarial networks (GAN) and influence the decision of the extractor.}
\label{fig:intro}
\end{figure*}

We follow the previous work~\cite{li2013joint,nguyen2016joint} to conduct the EE task: jointly detect the trigger words and arguments simultaneously and, specifically in our framework, with shared parameters in neural network structure, following the schema of Automatic Content Extraction (ACE)\footnote{https://en.wikipedia.org/wiki/Automatic\_content\_extraction} .

We model the joint EE task into a reinforcement learning problem, and we briefly introduce the terms of RL and their counterparts/equivalence in EE in Table~\ref{tab:term_mapping}.

In all, we describe the extractor's tasks as follows: Given a sentence, our extractor scans the sentence and determines the boundaries of entities and event triggers. In the meanwhile, the extractor also takes entities in the same sentence and determine the relations between triggers and entities -- \textit{a.k.a} argument roles. During the training epochs, generative adversarial networks estimate rewards which stimulate the extractor to pursue the most optimal policy (model).

Figure~\ref{fig:intro} demonstrates the diagram of the EE task where the extractor is emulating from the expert and the procedure of dynamic reward estimation on an example: the extractor commits a wrong action (labeling trigger ``\textit{resignation}'' as an \verb|End-Org| event) in early epochs, and the GAN expands the margins between the rewards issued on correct and wrong actions; these rewards guide the extractor to discover the correct action.

\section{Framework Overview}
\label{sec:framework}
\subsection{Sequence Labeling with Q-Learning}
\begin{figure*}[ht!]
  \includegraphics[width=0.95\textwidth]{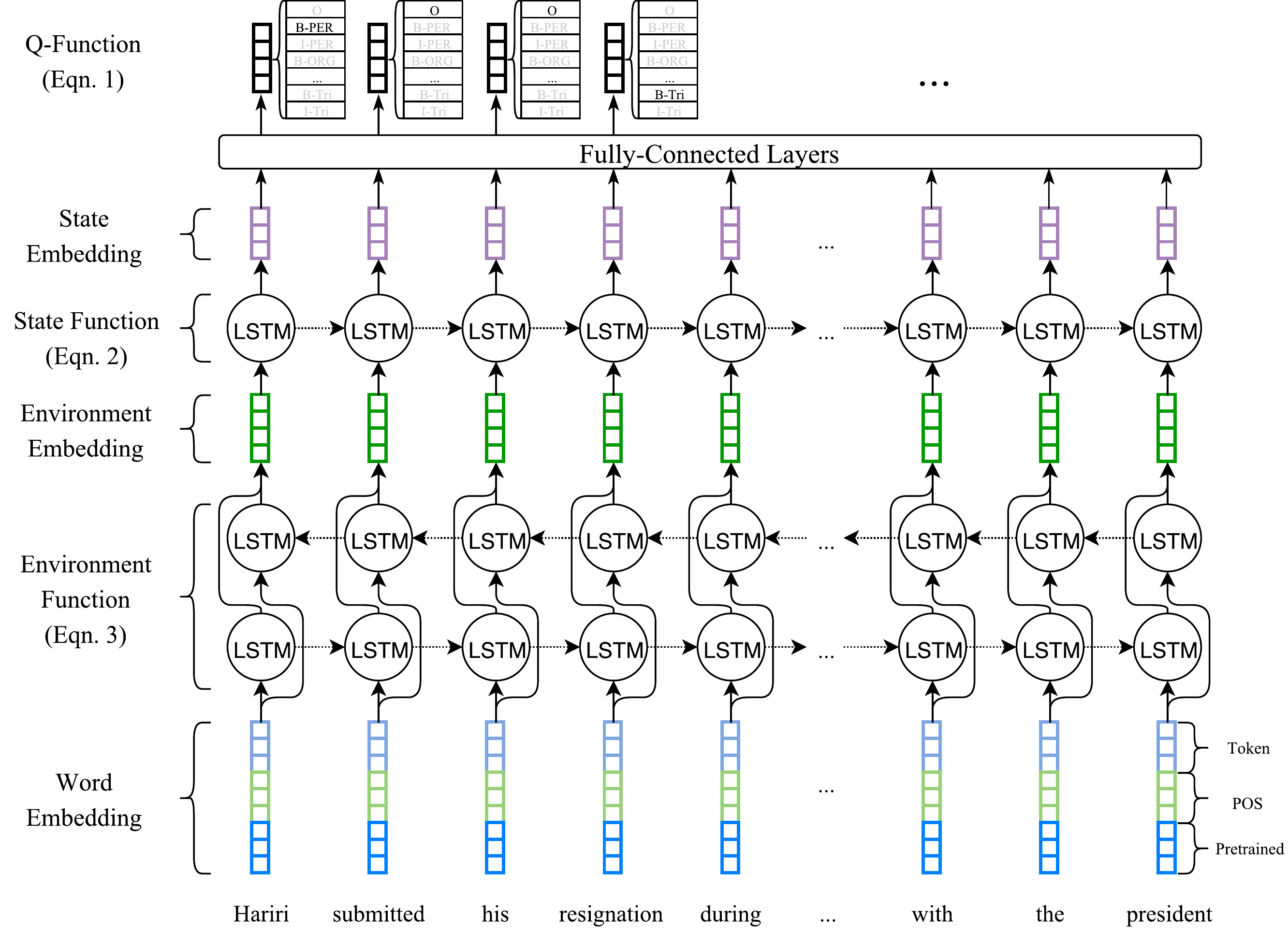}
  \caption{Structure overview of neural network for Section~\ref{sec:sequence_labeling}.  Details about word embeddings will be introduced in Section~\ref{sec:inputdropout}}
  \label{fig:lstm}
\end{figure*}
\label{sec:sequence_labeling}
Given a sentence $X = \{x_1, x_2, \ldots, x_n\}$ which consists of $n$ tokens, the extractor aims to search the correct action for the $t^{\text{th}}$  token $x_t$ in a given search space.
\begin{equation}
  \label{eqn:q_table}
  \hat{a}_t=\argmax_{a_t} Q_{sl}(s_t, a_t),
\end{equation}
where $a_t$ denotes the \textit{action} of labeling a token $x_t$ and can be considered as a BIO label\footnote{In this work, we use BIO style, e.g., ``B-PER'' indicates the token is the first token in a person entity, ``I-Trigger'' means that the token is inside an event trigger, and ``O'' denotes a ``None'' label.}.
$s_t$ denotes a state, $Q_{sl}(\cdot, \cdot)$ is a value function denoting the values of actions at state $s_t$.
\begin{equation}
\label{eqn:q_state}
  Q_{sl}(s_t, a_t) = \bm{f}_{sl}(s_t | s_1, \ldots, s_{t-1}, a_1, \ldots, a_{t-1}; \bm{\theta}_{sl})
\end{equation}
where $\bm{f}_{sl}(\cdot)$ is state function parametrized by $\bm{\theta}_{sl}$.
Equation~\ref{eqn:q_state} denotes that the extractor will determine its next action based on the previous states and actions.

Each state $s_t$ can be formulated as:
\begin{equation}
\label{eqn:state_from_env}
  s_t = \bm{f}_e(X, t;\bm{\theta}_e)
\end{equation}
where $\bm{f}_e(\cdot)$ is environment function.

Figure \ref{fig:lstm} demonstrates the neural network structure for sequence labeling. Equations \ref{eqn:state_from_env} implies that the extractor scans through the whole sentence, therefore, we use a Bi-LSTM~\cite{hochreiter1997long}, which extracts features -- or \textit{environment embeddings} in Figure~\ref{fig:lstm} -- from word embeddings in the whole sentence from both directions. We also utilize a single-direction LSTM to extract state embeddings as Equation~\ref{eqn:q_state}, followed by Fully-Connected layers to represent the Q function in Equation~\ref{eqn:q_table}. Note that the previous Q function values, which determine the previous actions $\{a_{t-1}, \ldots, a_1\}$, come from previous hidden output of the single-direction LSTM.

\begin{figure}[t!]
  \includegraphics[width=0.45\textwidth]{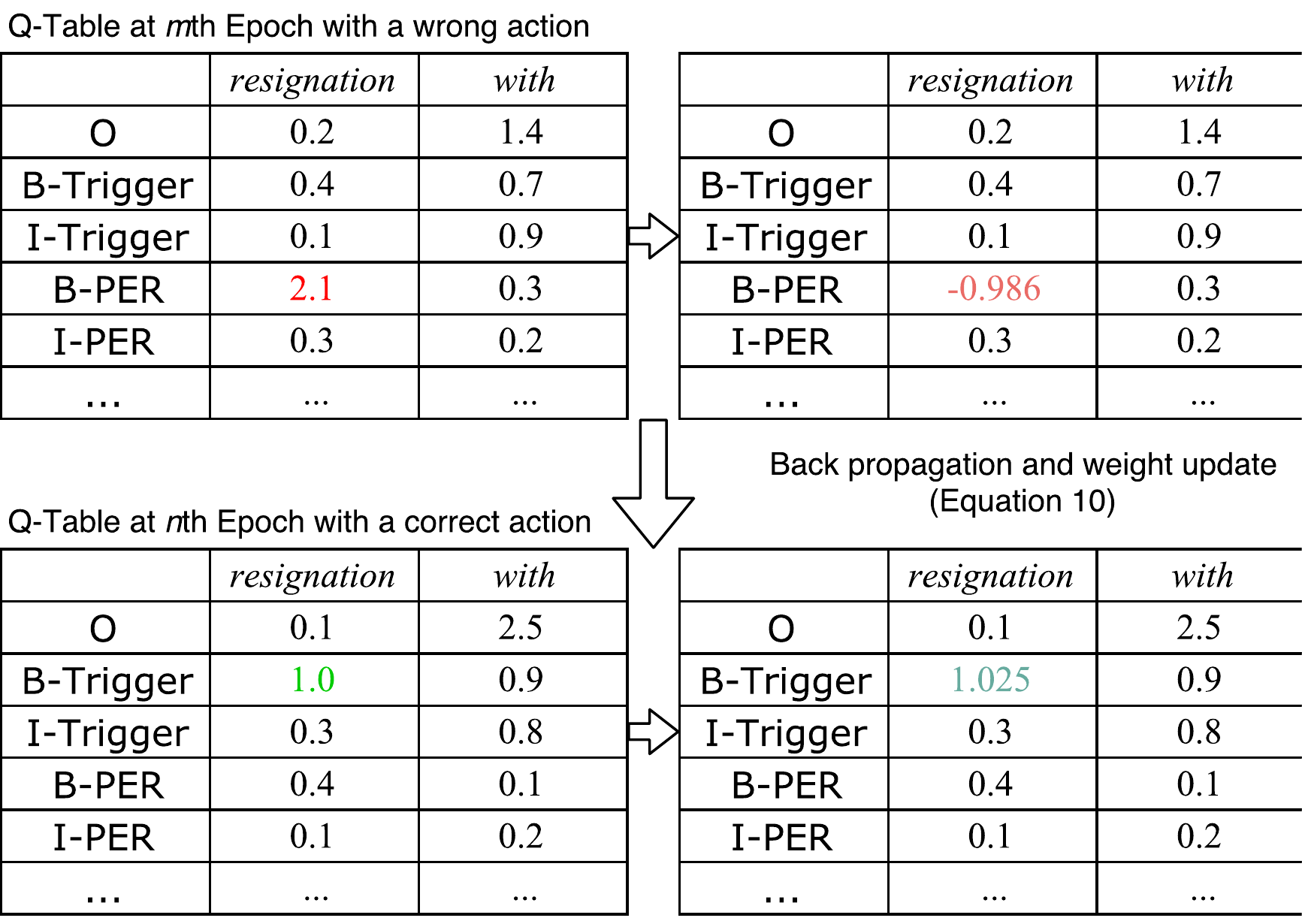}
  \caption{Update of Q function values with Equation~\ref{eqn:bellman_update}, with reward $r=\pm 1$ for correct/wrong actions and discount factor $\lambda=0.01$}
  \label{fig:q_learning}
\end{figure}
We utilize Q-Learning~\cite{watkins1992q} to train and optimize the values in Equation~\ref{eqn:q_state} to infer the most optimal policy $\pi^{*}$:
\begin{equation}
\label{eqn:q_function}
\pi^*(s) = \argmax_{a}Q(s,a)
\end{equation}
and we can also have
\begin{equation}
\label{eqn:policy_with_parameters}
\pi^*(s) = \argmax_{\bm{\theta}_{sl}, \bm{\theta}_{e}}\bm{f}_{sl}(\bm{f}_{e}(X; \bm{\theta}_{e});\bm{\theta}_{sl})
\end{equation}

At each step $t$, an action $a_t$ will be issued with a reward $r_t=R(s_t, a_t)$ with regard to the state $s_t$. In Q-learning, the value function $Q(\cdot, \cdot)$ is the expected value of the sum of future rewards.
\begin{equation}
\label{eqn:pi_expectation}
Q^{\pi}(s_t, a_t) = \mathbb{E}[R_t],
\end{equation}
where
\begin{equation}
R_t = \sum_{k=t}^n \gamma^{k-t} r_k.
\end{equation}
$\gamma$ is the \textit{discount factor} which determines the influence among current and future rewards.

To pursue the most optimal policy in Equation~\ref{eqn:q_function}, we use \textit{Bellman Equation}

\begin{equation}
\label{eqn:bellman_update}
Q^{\pi^{*}}_{sl}(s_t, a_t) = r_t + \gamma \max_{a_{t+1}}Q_{sl}(s_{t+1}, a_{t+1})
\end{equation}

We can iteratively update the Q functions by minimizing the mean squared error:
\begin{equation}
\label{eqn:q_object}
\mathbb{E}[(r_t + \gamma \max_{a_{t+1}}Q_{sl}(s_{t+1}, a_{t+1}) - Q(s_t, a_t))^2]
\end{equation}
and update the parameters in Equation~\ref{eqn:policy_with_parameters}. Since we have a neural network structure, we follow the optimization schema in Deep Q-Network~\cite{mnih2015human} to update the parameters $\bm{\theta}_{sl}$ and $\bm{\theta}_{e}$ and we use Adam Optimizer~\cite{kingma2014adam} to update the parameters.

Figure~\ref{fig:q_learning} demonstrates the process of updating the Q function.

\subsection{Event Extraction with Policy Gradient}
%

%
Similar to Equation~\ref{eqn:q_table}, the extractor selects the best action (event type) for a trigger, which is the ${t_{tr}}^{\text{th}}$ token in the sentence:
\begin{equation}
\label{eqn:trigger_table}
\hat{a}_{tr}=\argmax_{a_{tr}} Q_{tr}(s_{tr}, a_{tr}),
\end{equation}
and we also have
\begin{equation}
\label{eqn:trigger_state}
Q_{tr}(s_{tr}, a_{tr}) = \bm{f}_{tr}(s_{tr};\bm{\theta}_{tr}),  
\end{equation}
where $\bm{f}_{tr}(\cdot)$ is parametrized by $\bm{\theta}_{tr}$ and similarly we have 
\begin{equation}
\label{eqn:trigger_state_from_env}
s_{tr} = \bm{f}_e(X, t_{tr};\bm{\theta}_{e}),
\end{equation}
where $s_{tr}$ denotes a trigger state.

Finally the extractor explores actions on argument roles, given a pair of trigger and argument candidate $x_{t_{ar}}$.
\begin{equation}
\label{eqn:argument_table}
\hat{a}_{ar}=\argmax_{a_{ar}} Q_{ar}(s_{ar}, a_{ar}),
\end{equation}
and
\begin{equation}
  Q_{ar}(s_{ar}, a_{ar}) = \bm{f}_{ar}(s_{ar}; \bm{\theta}_{ar}).
\end{equation}
The state $s_{ar}$ should consider the trigger as well as the sequence label of the argument candidate, because there are constraints in argument role labeling, \textit{e.g.}, a \verb|PER| is never assigned with a \verb|Place| role. We have an argument state
\begin{equation}
  s_{ar} = <\bm{f}_e(X, t_{tr}; \bm{\theta}_{e}), \bm{f}_e(X, t_{ar}; \bm{\theta}_{e}), a_{t_{ar}}, \bm{d}>,
\end{equation}
where $\bm{d}$ denotes the dependency relation between the trigger token $x_{t_{tr}}$ and the argument token $x_{t_{ar}}$.

We utilize another RL algorithm, Policy Gradient~\cite{sutton2000policy} to determine the actions of selecting an event type for a trigger and assigning an argument role (or non-role) on the entity in the same sentence including the trigger.

Different from Equation~\ref{eqn:pi_expectation}, the Q function in \ref{eqn:trigger_table} and \ref{eqn:argument_table} is regarded as a probability distribution:
\begin{equation}
Q(s, a) = P(a|s),
\end{equation}
while they share the same goal of maximizing the expected value of sum of discounted rewards as Equation~\ref{eqn:pi_expectation}, and we follow
\begin{equation}
\label{eqn:pg_core}
\nabla_{\bm{\theta}_e, \bm{\theta}_{ev}}\mathbb{E}[R_t] =\mathbb{E}[\nabla_{\bm{\theta}_e, \bm{\theta}_{ev}}\log P(a|s)R_t]
\end{equation}
to directly optimize the policy, where $\bm{\theta}_{ev}$ denotes the parameters for trigger classification $\bm{\theta}_{tr}$ or argument role labeling  $\bm{\theta}_{ar}$.

To pursue Equation~\ref{eqn:pg_core}, we minimize
\begin{equation}
\label{eqn:pg_object}
-\log P(a|s) * R_t.
\end{equation}

From Equation~\ref{eqn:pg_object} we can acknowledge that, when the extractor commits a correct action, the reward encourages $P(a|s)$ to substantially increase; and when the action is wrong, the reward will be smaller or even negative, leading to a less increased or decreased $P(a|s)$.

\begin{figure}[t!]
  \includegraphics[width=0.45\textwidth]{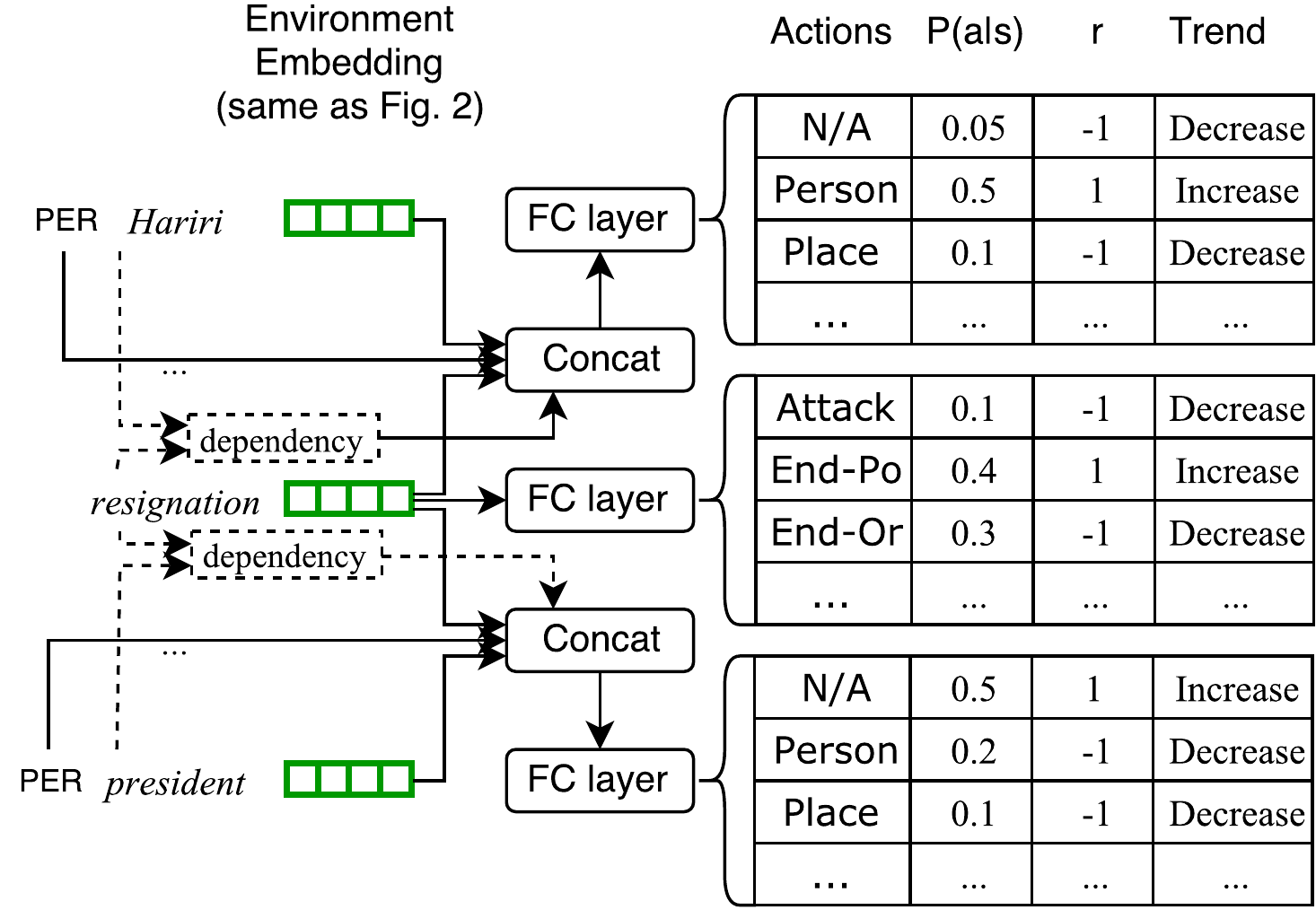}
  \caption{Network structure and updating process with policy gradient, the column of ``trend'' denotes the changes of $P(a|s)$ after policy gradient optimization/search in Equation~\ref{eqn:pg_object}.}
  \label{fig:policy_gradient}
\end{figure}

Figure~\ref{fig:policy_gradient} shows the network structure and process of update with policy gradient.

We also use Adam Optimizer to update the parameters. Note that Equations \ref{eqn:q_object} and \ref{eqn:pg_object} of all subtasks -- sequence labeling, event type classification and argument role labeling -- \textit{\textbf{jointly}} update the environment parameters $\bm{\theta}_e$ and input embeddings (see Section~\ref{sec:inputdropout}), the weights in the neural network are shared and information from later subtasks (event type and argument role) is able to provide feedback to the earlier sequence labeling.

\section{Generative Adversarial Imitation Learning}
By default the reward function $R(s, a)$ can be
\begin{equation}
  R(s, a) =
  \begin{cases}
    c_1 \text{ if the action is correct}\\
    c_2 \text{ if the action is wrong}
  \end{cases},
  \end{equation}
where $c_1$ and $c_2$ are constants and $c_1 > c_2$.
  
However, the labels on tokens, trigger type and argument roles in the EE task are complex, and the values of $R(s, a)$ are expected to be diverse and dynamic. For example, the extractor is expected to receive lower rewards if it labels a trigger token as entity than when it entirely misses the trigger token.
The RL frameworks applied on entity relation extraction~\cite{feng2017joint,zhang2017relation} provide a set of predefined reward values.
Such parameter settings are vulnerable and the tuning procedure is very expensive and inefficient.
Moreover, since the states in our framework are sampled from continuous space, we expect that the reward function should be continuous and differentiable as well.

Therefore, instead of adopting arbitrarily predefined, discrete yet risky reward values, we utilize inverse reinforcement learning, which estimates reward functions from the difference between the expert policy and the trained policy.

The IRL ensures that the highest rewards are issued to the expert unless the extractor commits exactly the same actions.
\begin{equation}
  \label{eqn:ensure_no_less_than}
  \mathbb{E}_{\pi_E}[R_t] \geq \mathbb{E}_{\pi_A}[R_t]
  \end{equation}

Such estimation and assessment on the differences among the expert and extractors can be considered as ``\textit{adversary}'', hence, we adopt Generative Adversarial Imitation Learning (GAIL)~\cite{NIPS2016_6391}, which is based on \textit{Generative adversarial network} (GAN)~\cite{goodfellow2014generative}. The core idea of GAN is establishing a generator to output fake data instances and a discriminator to distinguish them from real data. The output of the discriminator $D(\cdot)\in[0, 1]$ indicates the probability of the input data instance being real data, while in our framework, $D(\cdot)$ indicates the probability of the input actions and states being from the expert. We ensure: 
\begin{equation}
\label{eqn:ensure_D_no_less_than}
\mathbb{E}_{\pi_E}[D(s, a)] \geq \mathbb{E}_{\pi_A}[D(s, a)]
\end{equation}
According to Equation~\ref{eqn:ensure_no_less_than}, we can regard the output of the discriminator as an estimation on the reward function value $R(s, a)$.

In pursuit of Equation~\ref{eqn:ensure_D_no_less_than}, we utilize the object function as \cite{NIPS2016_6391}:
\begin{multline}
  \text{minimize} \max_{D \in (0,1)^{S\times A}} \mathbb{E}_{\pi_A}[\log D(s, a)]\\
  + \mathbb{E}_{\pi_E}[log(1-D(s, a))] - H(\pi),
\end{multline}
where $H(\cdot)$ indicates an entropy regularizer
\begin{equation}
  H(\pi) = - \sum_{s, a}p_{\pi_A}(a|s)\log p_{\pi_E}(a|s)
\end{equation}

Generally, the discriminator $D(\cdot)$ is a neural network activated by a sigmoid function bounded in $(0, 1)$ and we perform a linear transformation on the sigmoid function to bound the reward function value in $(-1, 1)$
\begin{equation}
  R(s, a) = 2 D(s, a) - 1
\end{equation}

\begin{figure}[t]
  \includegraphics[width=0.49\textwidth]{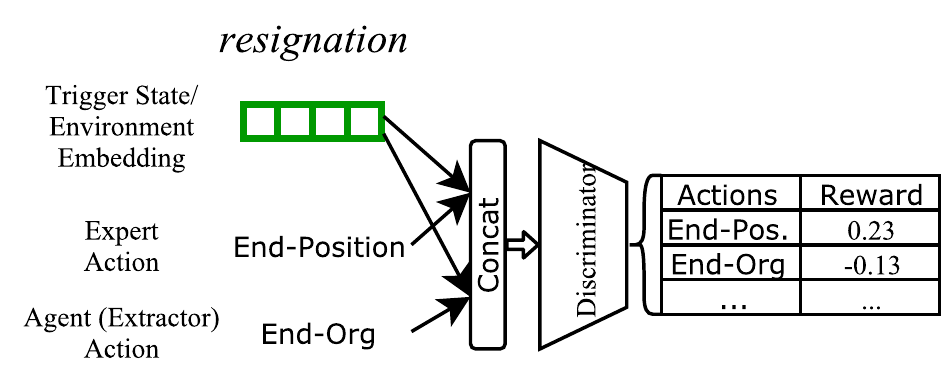}
  \caption{The example input (state, actions), structure and output (reward values with regard to actions) of discriminator.}
  \label{fig:gan}
\end{figure}

Figure~\ref{fig:gan} illustrates the input and output of the discriminator.

We apply Adam Optimizer to update the parameters in the discriminator. It is crucial to indicate that the discriminator networks (1 for sequence labeling, 1 for trigger classification and 33 for argument role detection) are independent from the LSTMs and optimization on those networks does not directly impact the parameters in the LSTMs. In contrast, LSTMs utilize the rewards -- the output of those discriminator networks -- to update themselves as presented in Section~\ref{sec:framework}.

\section{Input and Exploration}
\subsection{Inputs and Dropouts}
\label{sec:inputdropout}
We use the following word embedding techniques to represent tokens in the input sentence.
\begin{itemize}[noitemsep,nolistsep,leftmargin=*]
\item Token surface embeddings: for each unique token in the training set, we have a look-up dictionary which is randomly initialized and updated in the training phase. These embeddings represent the surface forms of the input tokens.
\item POS embeddings: We apply Part-of-Speech (POS) tagging~\cite{toutanova2003feature} on the sentences. The POS tags of the tokens also have a look-up dictionary similar to the one for token surfaces.
\item Pretrained embeddings: We also acquire embeddings trained from a large and publicly available corpus. These embeddings preserve semantic information of the tokens and they are not updated in the training phase.
\end{itemize}
We concatenate these embeddings and feed them into the Bi-LSTM network as Figure~\ref{fig:lstm} illustrates.

In order to robustly deal with noisy test data, \textit{e.g.}, instances with out-of-vocabulary (OOV) tokens, we utilize Dropout on the input data during the training phase. We intentionally set an ``OOV'' token, which holds an entry in the look-up dictionary. We randomly mask some known tokens in the training sentences with the ``OOV'' token. We also set an all-$0$ vector on pretrained embeddings of randomly selected tokens -- these tokens are selected independently regardless of OOV-masking. However, POS input will be preserved at all times, because they are typically closed sets of labels.

\subsection{Exploration Strategies}
\label{sec:exploration}
During the test procedure, the extractor will commit an action according to the Q-function values in Equation~\ref{eqn:q_table}, \ref{eqn:trigger_table} and \ref{eqn:argument_table}. In the training phase, we adopt $\epsilon$-greedy strategy: we set a probability threshold $\epsilon\in[0, 1)$ and uniformly pick up a number $\rho\in[0, 1]$
\begin{equation*}
  \hat{a} =
  \begin{cases}
    \argmax_{a} Q(s, a), \text{ if } \rho \geq \epsilon\\
    \text{Randomly pick up an action},\text{ if others}
  \end{cases}
\end{equation*}
In this way, the extractor is able to explore all possible actions in the search space, especially on some challenging instances or states.

\section{Experiments}
\subsection{Experiment Setup}
To evaluate the EE performance with our proposed approach, we utilize ACE2005 documents excluding informal documents from \verb|cts| and \verb|un| as mentioned in previous ACE EE work~\cite{li2013joint,nguyen2016joint}. We also follow their training, development and test splits and adopt the same criteria of the evaluation:
\begin{itemize}[noitemsep,nolistsep,leftmargin=*]
\item A trigger is correct if its event type and offsets find a match in the ground truth.
\item An argument is correctly labeled if its event type, offsets and role find a match in the ground truth.
\end{itemize}
We tune the parameters according to the F1 score of argument labeling and the parameters are presented in the Table~\ref{tab:parameter}.

For pretrained embeddings, we train a Word2Vec~\cite{mikolov2013distributed} model from English Wikipedia articles (January 1st, 2017), with all tokens preserved and a context window of 5 from both left and right.

\begin{table}[t]
\centering
\begin{tabular}{@{}ll@{}}
\toprule
Parameters              & Value \\ \midrule
Discount factor $\gamma$        & 0.01  \\
Hidden layer sizes (all FC and LSTM) & 256   \\
Token surface embedding dimension & 200   \\
PoS embedding dimension         & 100   \\
Pretrained embedding dimension & 200\\
Fixed rewards (RL baseline only) & $\pm 1$\\
Probability threshold $\epsilon$ & 0.1\\
Dropout rate & 0.05\\
Learning rate (for all Adam Optimizers) & 0.001\\
\bottomrule
\end{tabular}
\caption{Parameters in the experiments}
\label{tab:parameter}
\end{table}

\subsection{Results and Analysis}
\begin{table*}[ht]
  \centering
 \begin{footnotesize}
\begin{tabular}{lcccccccccccc}
\toprule
Tasks & \multicolumn{3}{c}{Trigger Identification}   & \multicolumn{3}{c}{Trigger Labeling} & \multicolumn{3}{c}{Argument Identification}   & \multicolumn{3}{c}{Role Labeling} \\
\cmidrule(l){2-4}\cmidrule(l){5-7}\cmidrule(l){8-10}\cmidrule(l){11-13}
  Metric                     & P & R & F1 & P & R & F1 & P & R & F1 & P & R & F1\\
  \midrule
  JointIE\cite{li2013joint} & \textbf{76.9} & 65.0 & 70.4 & 73.7 & 62.3 & 67.5 & \textbf{69.8} & 47.9 & 56.8 & 64.7 & 44.4 & 52.7 \\
  JRNN\cite{nguyen2016joint} & 68.5 & 75.7 & 71.9 & 66.0 & 73.0 & 69.3 & 61.4 & \textbf{64.2} & \textbf{62.8} & 54.2 & \textbf{56.7} & 55.4 \\
  RL (our approach) & 74.7 & 65.4 &  69.8 & 71.2 & 63.2 & 66.9 & 57.7 & 56.4 & 57.0 & 57.3   &  42.9    & 49.1  \\ 
  GAIL (our approach)  & 76.4 & \textbf{68.2} & \textbf{72.1} & \textbf{74.2} & \textbf{65.3} & \textbf{69.5} & 66.2 & 51.4 & 57.8 & \textbf{65.6}  & 48.7  & \textbf{55.9}  \\
\bottomrule
\end{tabular}
\end{footnotesize}
\caption{Performance comparison with State-of-the-Art frameworks on ground-truth entity annotation of ACE2005.}
\label{tab:gold}
\end{table*}

\begin{table*}[ht]
  \centering
 \begin{footnotesize}
\begin{tabular}{lcccccccccccc}
\toprule
Tasks & \multicolumn{3}{c}{Trigger Identification}   & \multicolumn{3}{c}{Trigger Labeling} & \multicolumn{3}{c}{Argument Identification}   & \multicolumn{3}{c}{Role Labeling} \\
\cmidrule(l){2-4}\cmidrule(l){5-7}\cmidrule(l){8-10}\cmidrule(l){11-13}
  Metric & P & R & F1 & P & R & F1 & P & R & F1 & P & R & F1\\
  \midrule
  JointIE & -- & -- & -- & 65.6 & \textbf{61.0} & 63.2 & -- & -- & -- & \textbf{60.5} & 39.6 & 47.9 \\
  GAIL(Our Approach) & 74.1 & 58.8 & 65.6 & \textbf{73.5} & 58.2 & \textbf{64.9}  & 58.9  & 46.1  & 51.7  & 56.1  & \textbf{43.9} & \textbf{49.8}  \\
\bottomrule
\end{tabular}
\end{footnotesize}
\caption{Performance on joint event extraction (no ground-truth annotation) comparison with State-of-the-Art framework from \cite{li2014constructing} on ACE2005.}
\label{tab:end_to_end}
\end{table*}

Table~\ref{tab:gold} demonstrates the comparison with state-of-the-art frameworks~\cite{li2013joint,nguyen2016joint} as well as our own baseline with fixed rewards. In these frameworks, the extractor has ground-truth annotation on entities.

\begin{figure}[t!]
  \centering
  \includegraphics[width=0.45\textwidth]{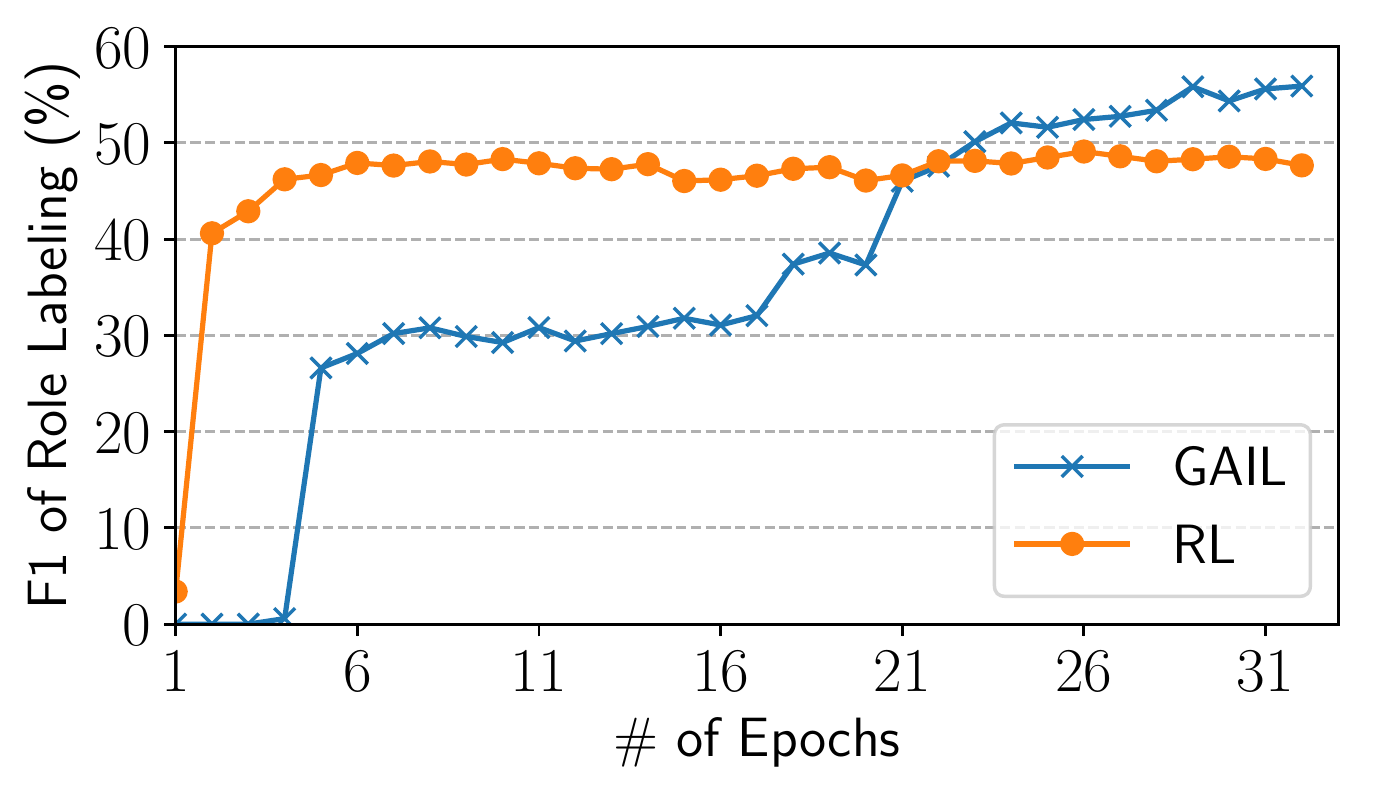}
  \caption{The F1-scores on argument labeling with epoch numbers.}
  \label{fig:pr_curve}
\end{figure}

We can conclude that the RL approach with fixed rewards does not outperform the other approaches. From Figure~\ref{fig:pr_curve} we see that at earlier epochs, the fixed reward approach achieves higher performance than GAIL approach: the extractor receiving fixed rewards acts with a more ``aggressive'' exploration strategy, but quickly converges and fluctuates around a performance limit; while the extractor receiving dynamic rewards from GAIL acts in a more ``conservative'' way: it starts with 3 epochs where every token is tagged as ``None'' and neither triggers nor arguments are detected. When the discriminator networks perform steadily, the extractor gradually acquires more optimal policies and commits more correct actions and finally outperforms the fixed reward approach.

\begin{figure}[t!]
  \centering
  \includegraphics[width=0.45\textwidth]{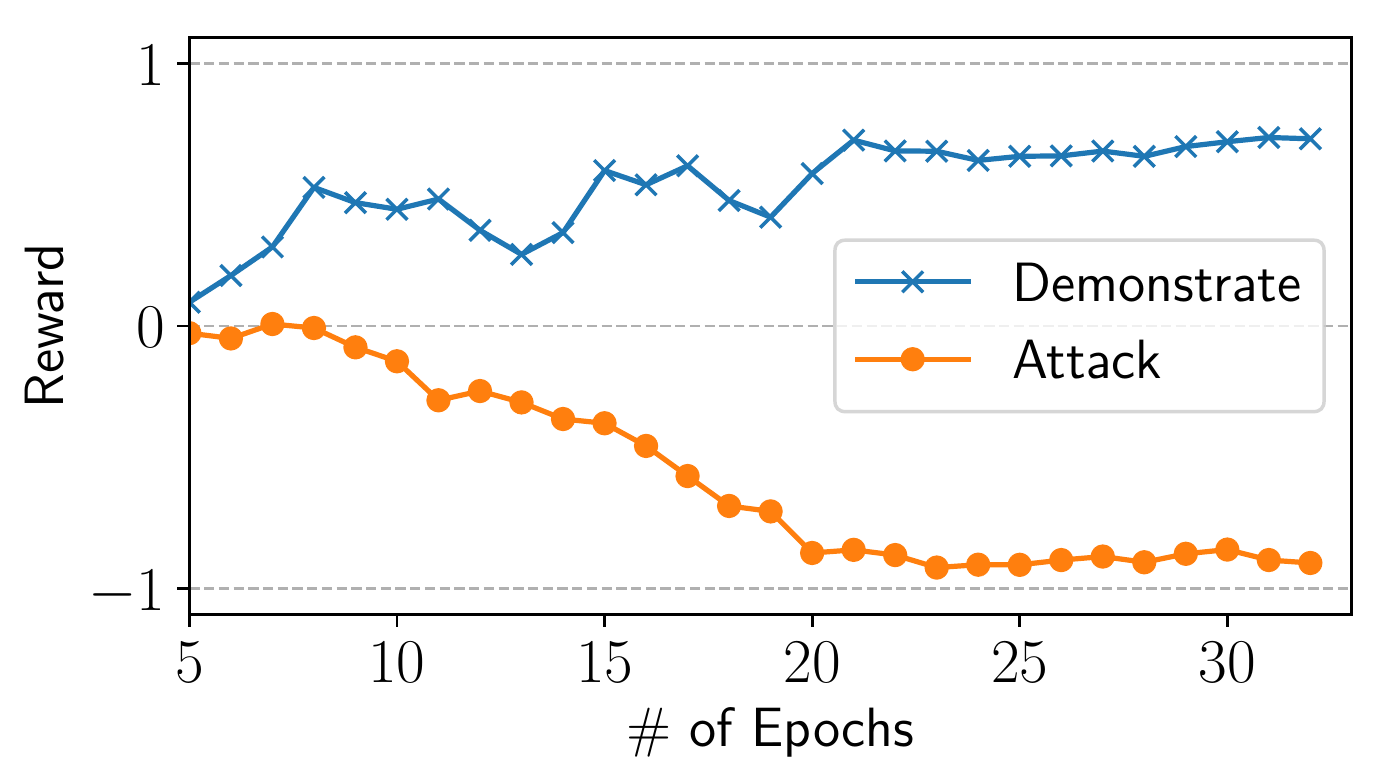}
  \caption{Change of rewards w.r.t. event type actions on the trigger ``\textit{\textbf{campaign}}'' .}
  \label{fig:reward_example}
\end{figure}

Regarding the comparison with other state-of-the-art frameworks, we can find that the performance of trigger labeling is better than the other frameworks. Figure~\ref{fig:reward_example} illustrates the curves of rewards with regard to two actions (\verb|Demonstrate| and \verb|Attack|) on the ``\textit{the disobedience \textbf{campaigns} began last week}'' example mentioned in Section~\ref{sec:intro}. In early epochs, the reward for tagging the word as \verb|Demonstrate| is slightly larger than the one for \verb|Attack|, but the extractor still commits \verb|Attack| action until the 23rd epoch, during which the margin expands. This observation meets our expectation -- repeating errors on the same instance will be considered as ``difficult'' and the reward margin between correct and wrong actions increases; and margin remains stable after the action for the instance is correct.

The performance of argument labeling is slightly better than the other frameworks. We still have advantage that we do not need excessive explicit feature engineering work in our proposed framework. 

In Table~\ref{tab:end_to_end}, we also compare our approach with another approach based on explicit feature engineering from \cite{li2014constructing} where the extractor is required to jointly extract entities, event triggers and argument roles using many linguistic resources. Our framework achieves better results than state-of-the-art, with a significant difference at 95\% confidence interval using Z-test.

Table~\ref{tab:trigger_vs_trigger_plus_entity} shows another scenario where the extractor is trained on triggers only in sequence labeling task and takes ground truth entity annotation as argument labeling for event extraction. The results drop drastically. Based on this observation, we can conclude that the extractor benefits from the parameter and embedding shared from training process with entity extraction, even thought it is not required to extract entities during test phase.

\begin{table}[t]
  \centering
 \begin{footnotesize}
\begin{tabular}{lcccccc}
\toprule
Tasks & \multicolumn{3}{c}{Trigger Labeling}   & \multicolumn{3}{c}{Role Labeling}  \\
\cmidrule(l){2-4}\cmidrule(l){5-7}
  Metric & P & R & F1 & P & R & F1 \\
  \midrule
  Trigger-only & 46.8 & 35.2 & 40.2 & 40.4 & 16.4 & 23.3  \\
  Trig.+Ent. & 74.2 & 65.3 & 69.5 & 65.6 & 48.7 & 55.9   \\
\bottomrule
\end{tabular}
\end{footnotesize}
\caption{Performance between extractors trained with different schema in sequence labeling.}
\label{tab:trigger_vs_trigger_plus_entity}
\end{table}


\section{Related Work}
We acknowledge that reinforcement learning has been recently applied to a few information extraction tasks. \cite{narasimhan2016improving} uses RL to acquire additional data; and \cite{feng2017joint,zhang2017relation} apply RL solely on entity relation detection and their entity detection is still based on a supervised method from \cite{huang2015bidirectional}, and these cascade frameworks do not share parameters. Our framework fully utilizes RL methods on an end-to-end procedure, and parameters and embeddings are shared and jointly trained across subtasks.

The term \textit{imitation learning} can also refer to \textit{behavior cloning} or \textit{data aggregation}~\cite{ross2011reduction,vlachos2013investigation,chang2015learning}. In these frameworks, the agent indirectly imitates the expert by recovering the cost function from the expert or estimating the mapping function from the data/state features to expert action, while the IRL frameworks (sometimes \textit{apprentice learning})~\cite{abbeel2004apprenticeship,syed2008apprenticeship,syed2008game,ziebart2008maximum,baram2017end} directly focus on reward functions.


\section{Conclusion}
In this paper we are apply imitation learning -- a reinforcement learning framework -- on event extraction.
The performance benefits from dynamic reward values and the proposed framework reduces the requirement of linguistic feature engineering.
Our current framework is built upon model-free RL approaches, indicating that the agent (extractor) observes states after committing actions. Our future work will focus on exploring model-based RL frameworks, where extractor will actively predict states before committing actions.

 \clearpage
\bibliography{acl2018}
\bibliographystyle{acl_natbib}
\end{document}